\documentclass[letterpaper, 10 pt, conference]{ieeeconf}  %

\IEEEoverridecommandlockouts  
\usepackage{kotex}
\usepackage{amsmath}
\usepackage{amssymb}
\usepackage{array}
\usepackage{caption}
\usepackage{graphicx}
\usepackage{multirow}
\usepackage{tabularx}
\usepackage{threeparttable}
\usepackage{color}
\usepackage{adjustbox}
\usepackage{gensymb}

\title{\LARGE \bf
Object-based SLAM utilizing unambiguous pose parameters \\considering general symmetry types
}

\author{Taekbeom Lee$^{\dagger}$, Youngseok Jang$^{\dagger}$, and H. Jin Kim
\thanks{Taekbeom Lee and H. Jin Kim are with the Aerospace Engineering Department, Seoul National University, South Korea
        {\tt\small ltb1128@snu.ac.kr, hjinkim@snu.ac.kr}}%
\thanks{Youngseok Jang is with the Mechanical and Aerospace Engineering Department, Seoul National University, South Korea
        {\tt\small duscjs59@gmail.com}}%
\thanks{$^{\dagger}$ Taekbeom Lee and Youngseok Jang are contributed equally to this manuscript.}%
\thanks{{This research was supported by Unmanned Vehicles Core Technology Research and Development Program through the National Research  Foundation of Korea(NRF) and Unmanned Vehicle Advanced Research Center(UVARC) funded by the Ministry of Science and ICT, the Republic of Korea(NRF-2020M3C1C1A01086411)}}%
}

\begin{document}

\maketitle
\thispagestyle{empty}
\pagestyle{empty}

\begin{abstract}

Existence of symmetric objects, whose observation at different viewpoints can be identical, can deteriorate the performance of  simultaneous localization and mapping (SLAM). This work proposes a system for robustly optimizing the pose of cameras and objects even in the presence of symmetric objects. We classify objects into three categories depending on their symmetry characteristics, which is efficient and effective in that it allows to deal with general objects and the objects in the same category can be associated with the same type of ambiguity. Then we extract only the unambiguous parameters corresponding to each category and use them in data association and joint optimization of the camera and object pose. The proposed approach provides significant robustness to the SLAM performance by removing the ambiguous parameters and utilizing as much useful geometric information as possible. Comparison with baseline algorithms confirms the superior performance of the proposed system in terms of object tracking and pose estimation, even in challenging scenarios where the baseline fails. 

\end{abstract}

\section{INTRODUCTION}

Simultaneous localization and mapping (SLAM), one of the core technologies for autonomous driving, is a technology that reconstructs the environment around the robot and estimates the robot position in the reconstructed map. 
Despite a significant progress in precise localization using geometrical information of the surrounding environment, it is still difficult for a SLAM system to achieve advanced tasks based on human interaction and scene understanding. To overcome the problem, semantic SLAM that uses semantic information in SLAM has been in using deep learning techniques such as recently developed instance segmentation \cite{he2017mask}, \cite{du20203dcfs}, \cite{xu2019mid} and 3d object detection \cite{rukhovich2021fcaf3d}, \cite{ding2019votenet}, \cite{pon2020object}. 

Object-based SLAM \cite{yang2019cubeslam}, \cite{salas2013slam++} is a branch of semantic SLAM that reconstructs an object-based map and provides high-level information. Object-based SLAM simultaneously estimates the location of semantic object inferred from the network while performing feature-based localization used in existing SLAM systems, and enables to estimate camera poses even in featureless environments. In addition, a more expressive map can be constructed by expressing the pose, type, and shape of the semantic objects in the map.

However, if the shape of the object itself is symmetric or if the observed shape is symmetric due to occlusion, object-based SLAM can suffer from significant error in the estimation process of ego motion and pose of the object. Since symmetric objects may have the same observation at different viewpoints, data association or motion estimation may fail.

\begin{figure}[t] 
\centering
\includegraphics[width = 1\linewidth]{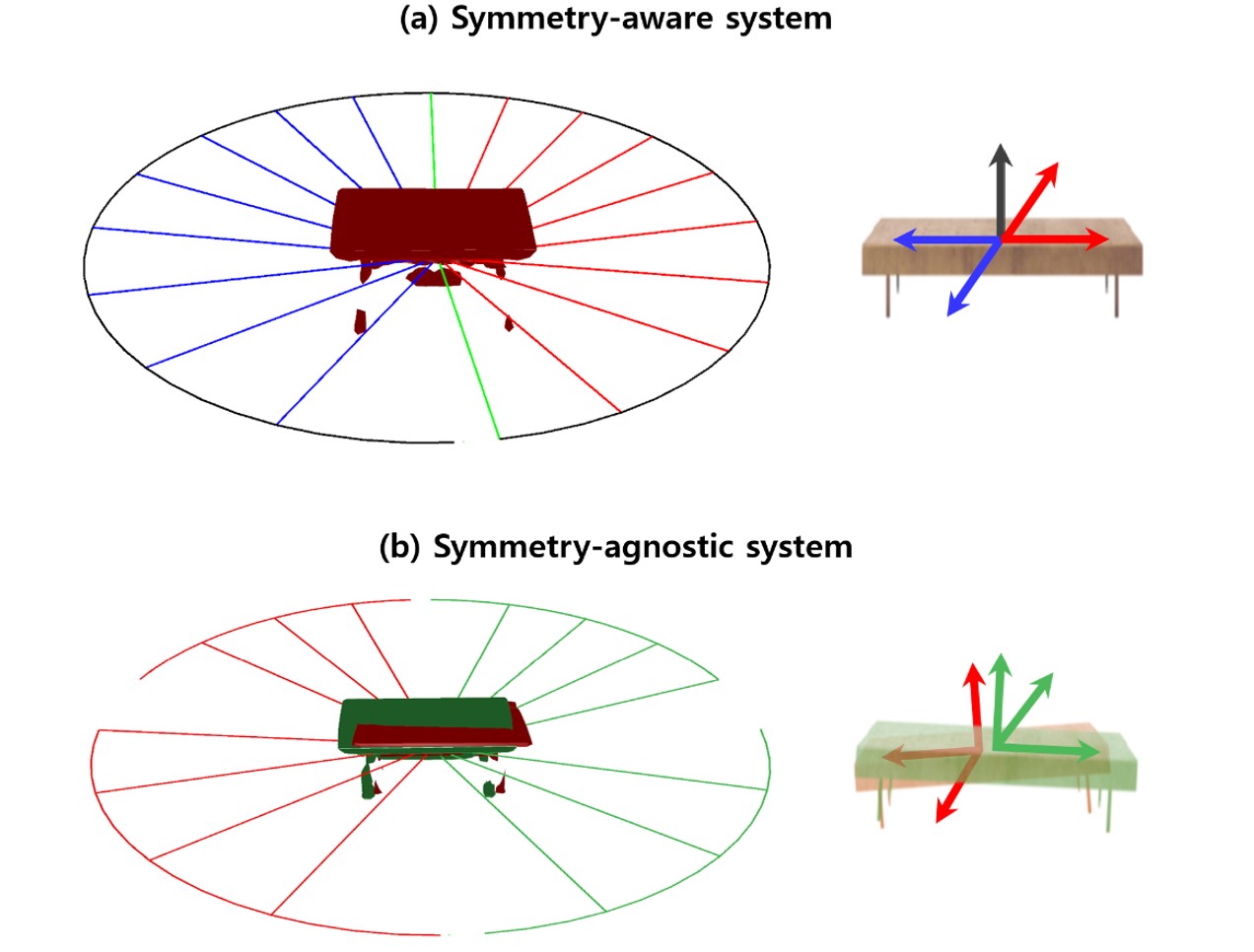}
\caption{{\textbf{} 
Illustration of object tracking results using the data obtained surrounding a rectangular table. Although only a single object is observed, the symmetry-agnostic approach in (b) may generate multiple map objects and fail object tracking. On the other hand, the symmetry-aware approach in (a) recognizes the discrete symmetry type of the object and succeed in object tracking using all the reliable measurements. Such improvement of (a) makes the object-based SLAM robust, since a large number of constraints obtained from long tracking of objects are useful for backend optimization.}}
\vspace{-3mm}
\label{fig: thumbnail}
\end{figure}

We propose a system for robustly optimizing the pose of cameras and objects even in the presence of symmetric objects. The 3d detection module \cite{rukhovich2021fcaf3d} is modified to obtain a local observation that predicts multiple poses for each object, similar to \cite{fu2021multi}. We classify object types into asymmetry, discrete, and continuous symmetry based on the local observation of the observed objects. 
By extracting a  parameter corresponding to the symmetry type from the pose of object, only the robust parameter is used as a constraint to jointly optimize  the camera pose and object pose.
Therefore, the proposed SLAM system has the advantage of robustness by using as much useful geometric information as possible even when symmetric objects are observed, while directly using 3d detection networks with many existing research and dataset. In summary, the contributions of the paper are as follows:

\begin{itemize}
    \item We design a symmetry and pose ambiguity aware object SLAM system which fully utilizes information from multiple pose hypotheses of objects to jointly optimize camera pose and globally consistent object pose.
    
    \item We propose a method to extract reliable information from ambiguous pose of symmetric objects. We categorize symmetry types of general objects and distinguish pose parameters into ambiguous one and unambiguous ones. The extracted unambiguous parameters of each symmetry type are used in the proposed object association and optimization modules.
    
    \item We use multiple hypotheses 3d detection network as observation module of our system, which can be easily edited from existing networks and changed to better networks in the future.
\end{itemize}	

The remainder of this paper is organized as follows. Section \ref{related work} reviews related literature. The overview and core concepts of the system are provided in Section \ref{system overview}. Detailed methods for applying core concepts to systems are described in Sections \ref{categorization} and \ref{joint BA}.
In Section \ref{experiments and results}, experiments in simulation and public dataset are presented. Finally, conclusions are provided in Section \ref{conclusions}. 

\begin{figure*}[t]
\centering
\includegraphics[width = 1\linewidth]{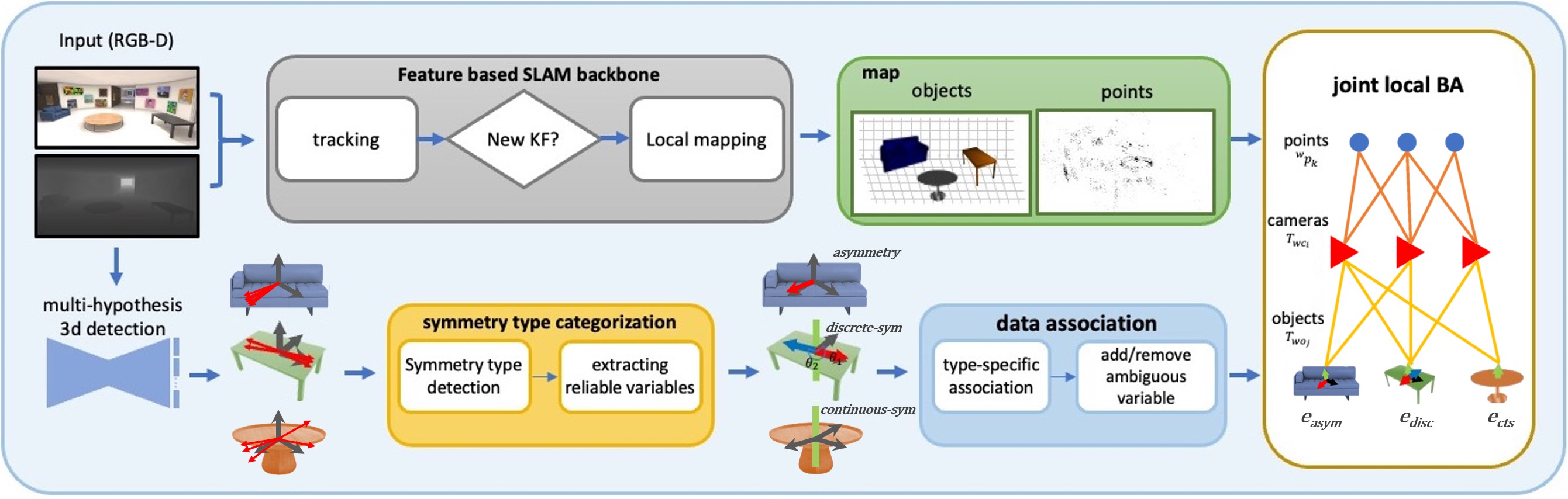}
\caption{\textbf{} The overview of the proposed object-based SLAM system.}
\vspace{-3mm}
\label{fig:pipeline}
\end{figure*}

\section{RELATED WORK}
\label{related work}
This section reviews related studies on object-based SLAM and symmetric-aware object pose estimation.

\subsection{Object-based SLAM}
Object-based SLAM attempts to build a robust SLAM system by simultaneously optimizing camera poses, feature positions, and poses of objects from semantic information.
To express the shape and pose of an object, \cite{salas2013slam++}, \cite{yang2019cubeslam}, and \cite{nicholson2018quadricslam} use prior object model, cuboid, and ellipsoid, respectively. And \cite{sucar2020nodeslam}, \cite{wang2021dsp} represent the shape and pose of an object using category specific embedding method. Although they showed that joint optimization could increase the robustness of both camera and object pose estimation, they did not consider the presence of ambiguous detection due to symmetric objects or occlusion.

\subsection{Symmetry-aware object pose estimation}
\subsubsection{Single view}
\cite{corona2018pose} proposes a network that can determine whether a shape has symmetry using an object's CAD model, and \cite{manhardt2019explaining} predicts multiple candidate poses for the detected object and analyzes ambiguity that may be caused by occlusion as well as ambiguity by the shape of an object. However, they estimated object pose only from a single view and did not treat multi-view cases such as SLAM.

\subsubsection{Multiple view}
Recently, studies considering the pose-ambiguity of objects have been proposed in object-based SLAM. PoseRBPF \cite{deng2021poserbpf} estimates the pose distribution of asymmetric and symmetric objects using the rao-blackwellized particle filter. However, it requires a predefined codebook for the pose of the object model. \cite{merrill2022symmetry} is the object-based SLAM system that simultaneously optimizes camera poses and object poses using the projection of reconstructed 3d keypoints as a prior. However, every time they encounter on a new dataset, they have to label the prior. \cite{zhen2022unified} expresses geometrical primitives in an unified, decomposed quadric form and explicitly deals with the degenerative case due to a symmetric shape.
However, the degenerative cases considered in \cite{zhen2022unified} are limited to a few quadric types, so error can be induced when fitting arbitrary shapes to quadric.
\cite{fu2021multi} uses a single neural network to predict poses as multi-hypotheses and optimizes them through a max-mixture\cite{olson2013inference} model. Since the symmetric object has ambiguous pose parameters, such as the rotation angle for an axis of symmetry, pose estimation should be performed by considering only the unambiguous elements. However, error can occur because \cite{fu2021multi} directly uses hypotheses containing even ambiguous parameters for pose estimation.

We propose a robust SLAM system using existing well-developed 3d detection modules while extracting only available geometric elements from symmetric objects and using them as optimization constraints.

\section{Proposed system}
\label{system overview}
\subsection{Symmetry types}
\label{Symmetry types}
The key idea of the proposed system is to propose a criterion that can effectively classify general objects using only three symmetry types and to define different pose representations suitable for each symmetry type in order to fully utilize the geometry information available in the symmetric object.

In general, objects can be asymmetric or symmetric. The asymmetric object can be inferred to have a consistent object pose in the 3d detection module, no matter which viewpoint it is observed from, but the symmetric object cannot. In addition, we classify symmetry as discrete and continuous types. Discrete symmetric objects refer to reflection symmetric objects that can have a finite number of poses as the observed viewpoints change, and continuous symmetric objects refer to objects with an infinite number of poses based on an axis of symmetry, such as a circular table. Objects which are classified into the following three types are represented in different ways to reflect all possible unambiguous pose parameters:
\begin{itemize}
    \item \textbf{Asymmetry:}
    All detection results of a asymmetric object can be used to optimize a unique pose since the unique pose can be determined when viewpoint changes. Accordingly, the pose can be represented by 6 degrees of freedom (DoF) which is commonly used.
    \item \textbf{Discrete symmetry:}
    Objects with multiple planes of symmetry have as many poses supporting the same shape as the number of planes of symmetry. We express the pose assuming that the symmetrical planes of most discrete symmetric objects present have one intersection line. The intersection of symmetric planes is defined as the axis of symmetry, and the rotation angle based on the axis is defined as a symmetric angle. Accordingly, the position and the axis of symmetry are shared with the reflected poses of a discrete symmetric object, and only the symmetric angle can be expressed differently. In other words, the pose of a discrete symmetric object is defined as five shared parameters for position and axis of symmetry and unshared parameters for all the symmetry angles.
    \item \textbf{Continuous symmetry:}
    For objects such as round tables, there is an infinite number of symmetry planes, so there is an infinite number of object poses that support the same shape. For continuous symmetric objects, it is assumed that planes of symmetry have a single intersection, as in the case of discrete symmetry. Therefore, we express the pose of continuous symmetric objects only by position and axis of symmetry after removing the symmetric angle by classifying it as an ambiguous parameter.
\end{itemize}

\subsection{Pipeline}
The entire pipeline is described in Fig. \ref{fig:pipeline}, and the tracking and object mapping modules are designed based on DSP-SLAM \cite{wang2021dsp}. When a new keyframe is selected, the multi-hypothesis detection network uses an rgb-d image to detect objects, and the detected objects infer observable multiple pose hypotheses (Section \ref{Multi-hypothesis 3d object detection}). The categorization module determines the symmetry type of the detected object based on the distribution of multiple pose hypotheses. We extract the axis of symmetry for symmetric objects and cluster the poses with similar symmetry angles for discrete symmetry objects (Section \ref{Symmetry type categorization of detection}). We associate the categorized detection with map objects using the class and pose except for ambiguous parameters. For discrete symmetric objects where a new symmetry angle is observed, we add a non-shared parameter (Section \ref{Data association of objects}). The camera pose, map point, and map object are jointly optimized at the backend. The constraint between the camera pose and map object pose is formed differently for each map object type during optimization (Section \ref{joint BA}).

\section{CATEGORIZED DETECTION AND OBJECTS}
\label{categorization}
\subsection{Multi-hypothesis 3d object detection}
\label{Multi-hypothesis 3d object detection}
We modify the 3d object detection module so that it can infer multiple pose hypotheses. Since the distribution of inferred multiple hypotheses influences the determination of the object's symmetry type, it should be modified so that the multi-hypothesis 3d object detection module can cover all the poses that the object can have. \cite{rupprecht2017learning} extends the single-loss single-output system to have multiple outputs, calculating loss \cite{guzman2012multiple} using only the hypothesis that succeeded in the most accurate inference among multiple hypotheses in the training process. Unlike the method of using the average loss of multiple outputs, this can increase multi-hypothesis diversity because each hypothesis is randomly selected and learned individually. We also modified the 3d object detection module \cite{rukhovich2021fcaf3d} in the same way as \cite{rupprecht2017learning}, allowing a source for the object's symmetry type to be inherent in multiple hypotheses.

\subsection{Symmetry type categorization of detection}
\label{Symmetry type categorization of detection}
\begin{figure}[t]
\centering
\includegraphics[width = 1\linewidth]{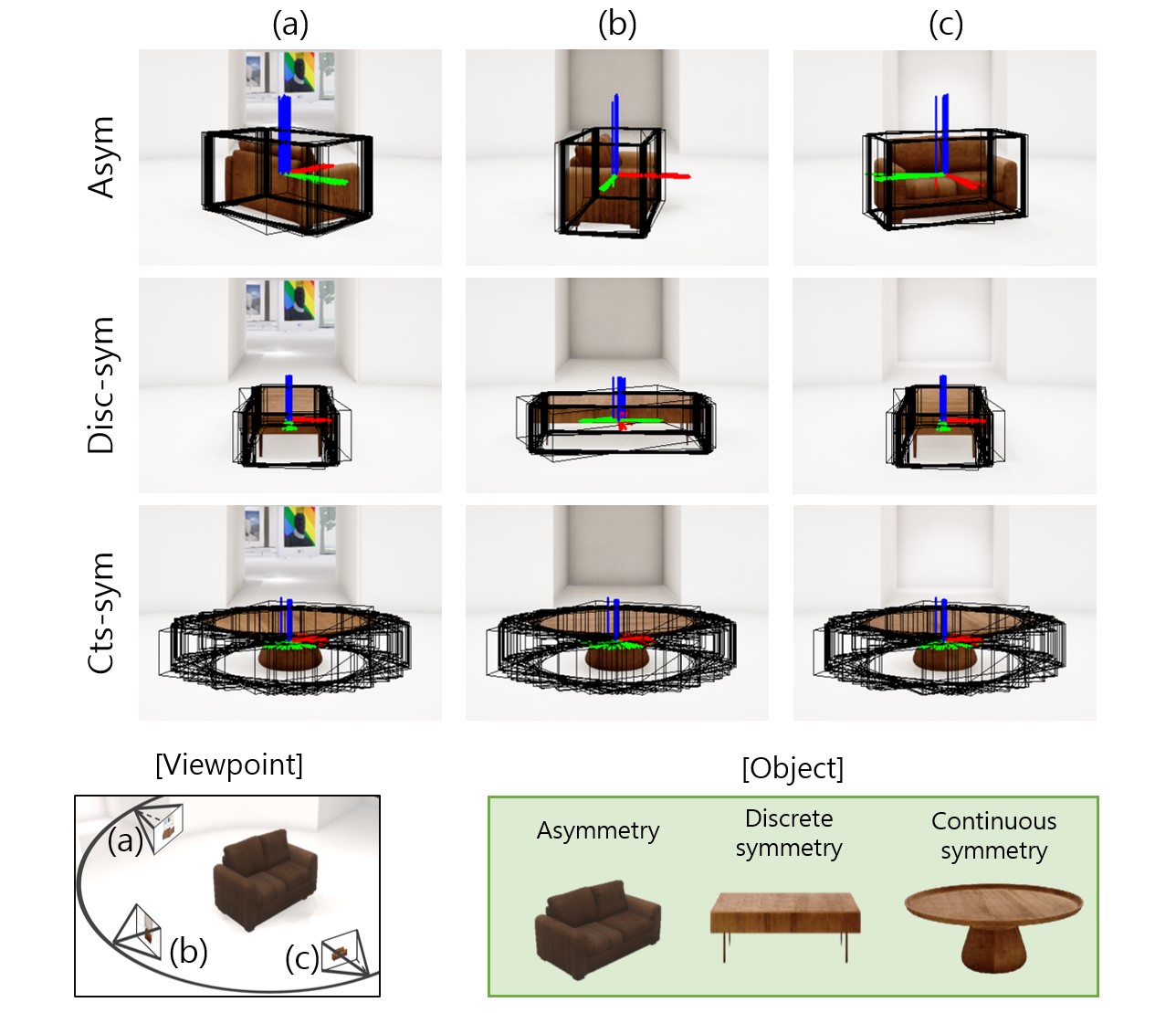}
\caption{\textbf{} Multi-hypothesis results for each symmetry type. Objects of three types are observed from different viewpoints (a), (b), and (c).}
\vspace{-5mm}
\label{fig:method_detection}
\end{figure}

As shown in Fig. \ref{fig:method_detection}, since the multi-hypothesis detection implies the ambiguity of the object's pose, we can classify the object's symmetry type and extract the type-specific pose parameters only from the detection result without prior information.

Unlike the symmetric object, the multiple pose hypotheses of the asymmetric object are similar to single pose. In other words, the distribution of pose hypotheses of the asymmetric object is very close to unimodal, and the variance is much smaller than that of the symmetrical object.

For discrete and continuous symmetry types, we perform an additional classification process because the normalized singular values of multiple hypotheses are large in both cases. As mentioned in Section \ref{Symmetry types}, both types have certain axis of symmetry, and the axis of symmetry $l$ is obtained as follows:
\begin{align}
    l^{*}=\underset{{l\in\mathbb{R}^{3}}}{\text{argmax}}\left\|[\overline{\omega}_{o_{1}o_{2}},\overline{\omega}_{o_{1}o_{3}},\cdots,\overline{\omega}_{o_{1}o_{N}}]^{T} \cdot l \right\|_{2}~,
\end{align}
\begin{align*}
    \text{where} \quad \overline{\omega}_{o_{1}o_{i}} = \frac{\omega_{o_{1}o_{i}}}{\left\|\omega_{o_{1}o_{i}} \right\|_{2}}, \; \omega_{o_{1}o_{i}} = \log_{\text{so(3)}}(R_{co_{1}}^{T}R_{co_{i}}) \in \mathbb{R}^{3}.
\end{align*}
$N$ and $\overline{\omega}$ are the number of hypotheses and rotation axis, respectively. And, symmetry angles are computed by  $\theta_{o_{1}o_{i}} = \left\|\omega_{o_{1}o_{i}} \right\|_{2}$. After clustering the multiple hypotheses using the DBSCAN algorithm \cite{ester1996density}, the variance between the representative $\theta$ of each cluster is compared. As illustrated in Fig. \ref{fig:method_detection}, compared to the continuous symmetric object which corresponds to a continuous set of  $\theta$, the variance among clusters of discrete symmetric object is very large. Representative $\theta$ values of rectangular table form two clusters with about 180$\degree$ difference, while the round table has many similar representative $\theta$ clusters.

Then, reliable parameters of pose can be extracted for each classified symmetry type. First, the asymmetric object can use 6 DoF from detection results, and the continuous symmetric object can use position and axis of symmetry. In discrete symmetric objects, parameters are position, axis of symmetry, and representative symmetry angles of the clusters.

\subsection{Data association of objects}
\label{Data association of objects}
The categorized detection results are associated with the previously generated map objects.
First, the detection results, which represent relative transformation between the camera and the object, are warped in world coordinate using tracked camera pose $T_{wc}$. Then, object matching is performed by comparing the unambiguous parameters of the detected object and map objects considering the symmetry type. Due to inaccurate detection or partial observation, the same object can be categorized by different symmetry types in different views. We address the misclassification by following association strategies. If there is a discrete or continuous symmetry type in the two comparison groups, the position and axis of symmetry (5 DoF) parameters are compared; otherwise, the distance between the 6 DoF parameters is computed. If the distance is small enough and the class is the same, matching between the objects is successful.

After object matching is performed, the symmetric angles of the detected discrete symmetric object are associated with the nearest symmetric angle of the matched map object. The unmatched angles are added to the map object as a new symmetric angle.

Discrete symmetric object can be occasionally classified into asymmetry type at a specific viewpoint, as can be seen from Fig. \ref{fig:method_detection} (a) and (c) of the the discrete case. Therefore, even if 6 DoF matching between asymmetry objects fails, if 5 DoF matching is successful, we change the map object to the discrete symmetry type to express the pose with position, an axis of symmetry, and symmetric angles. Object association using unambiguous parameters for each symmetry type alleviates mismatching caused by ambiguous parameters. Therefore, the proposed system can track objects for a long time, which acts as an important source in joint optimization.

The detection result that failed to match is registered as a new map object after the shape reconstruction using deep-sdf \cite{park2019deepsdf} similar to DSP-SLAM.

\section{JOINT OPTIMIZATION}
\label{joint BA}
We perform joint optimization by modifying the existing SLAM optimization problem using the categorized map objects and associated detection results.
\begin{align}
C^{*}, O^{*}, P^{*} = \underset{{C,O,P}}{\text{argmin}} ~ E_{reproj}(C, P)+E_{obj}(C, O)~,
\end{align}
where C, O, and P refer to the camera and object pose and map point included in the optimized window size, respectively. $E_{reproj}$ denotes the reprojection error, and $E_{obj}$ denotes the pose error between the map object and the camera. The joint optimization is solved by the Levenberg-Marquardt method through g2o\cite{grisetti2011g2o} solver.

$E_{obj}$ is designed only with unambiguous parameters in consideration of the symmetry type of map object, which can produce a more robust solution than existing methods that utilize all pose parameters with an ambiguous parameter as a constraint.

\subsubsection{Asymmetry}
The results of asymmetry map object and associated multiple hypotheses are not linked by multiple edges but by a single edge using a max-mixture model, which selects the hypothesis with the lowest error at each iteration of optimization, similar to \cite{fu2021multi}.
\begin{align}
    e_{\text{asym}}(T_{wc},T_{wo}) = \min_{j\in{[1,N]}}\log_{\text{se(3)}}{({T_{co}^{j}} \cdot T_{wo}^{-1} \cdot T_{wc})}~,
\end{align}
where $T_{wc}$ and $T_{wo}$, which mean the pose of the camera and object on the world, are optimization variables, $T_{co}^j$ represents the $j$-th hypothesis, and $N$ is the total number of hypotheses.

\subsubsection{Discrete symmetry}
Discrete symmetric map objects have position and axis of symmetry as shared parameters and $M$ symmetric angles as non-shared parameters. Each hypothesis is associated with one of symmetry angles of the matched map object, and edges are formed as many as the number of associated symmetric angles, $m$. Each edge is modeled with the max-mixture like the asymmetry case.
\begin{align}
    e_{\text{disc}}(T_{wc},T_{wo_i}) = \min_{j_i\in{[1,N_i]}}\log_{\text{SE(3)}}{({T_{co}^{j_i}} \cdot T_{wo_i}^{-1} \cdot T_{wc})}~,
\end{align}
\begin{align*}
    \text{where} ~ T_{wo_i} = \begin{bmatrix}
    \exp{(\theta_{wo_i}\cdot\overline{\omega}_{wo})} & & t_{wo} \\ \text{0}_{1\times3} & 1
    \end{bmatrix}, ~ i \in{[1,m]}~.
\end{align*}
$N_i$ refers to the number of hypotheses associated with the $i$-th symmetric angle, and $T_{wo_i}$ means the $i$-th symmetric pose constructed by position and axis of symmetry which are shared parameters and the $i$-th symmetric angle. Furthermore, for unconstrained parameterization in optimization, the axis of symmetry is used by the following expression: 
$\overline{\omega}_{wo} = \text{f}(\phi_{wo}, \psi_{wo})$, where $\phi_{wo}$, $\psi_{wo}$ and $\text{f}(\cdot)$ are polar angle, azimuth in spherical coordinates, and the transformation function from $(\phi, \psi)$ to $\overline{\omega}$, respectively.

\subsubsection{Continuous symmetry}
The continuous symmetric map object has only position and axis of symmetry as unambiguous parameters. The axis of symmetry ($\overline{\omega}_{co}$) extracted from the rotation part and individual position parts of multiple hypotheses are used to formulate the following:
\begin{align}
    e_{\text{cts}}(T_{wc},T_{wo}) = e_{\text{trans}} + \gamma \cdot e_{\text{axis}}~,
\end{align}
\begin{align*}
\begin{split}
    \text{where}~
    e_{\text{trans}} &= \min_{j\in{[1,N]}} \| t_{co}^j - R_{wc}^{\text{T}}(t_{wc}-t_{wo}) \|_2~,\\
    e_{\text{axis}} &= \| \text{f}^{-1}(R_{wc} \cdot \overline{\omega}_{co}) - [\phi_{wo}, \psi_{wo}]^{\text{T}} \|_2~.
\end{split}
\end{align*}
$e_{\text{trans}}$ and $e_{\text{axis}}$ are max-mixture based translation error and error related to axis of symmetry, respectively. $\gamma$ is the constant weight for balancing two error terms.

\begin{figure}[t]
\centering
\includegraphics[width = 1\linewidth]{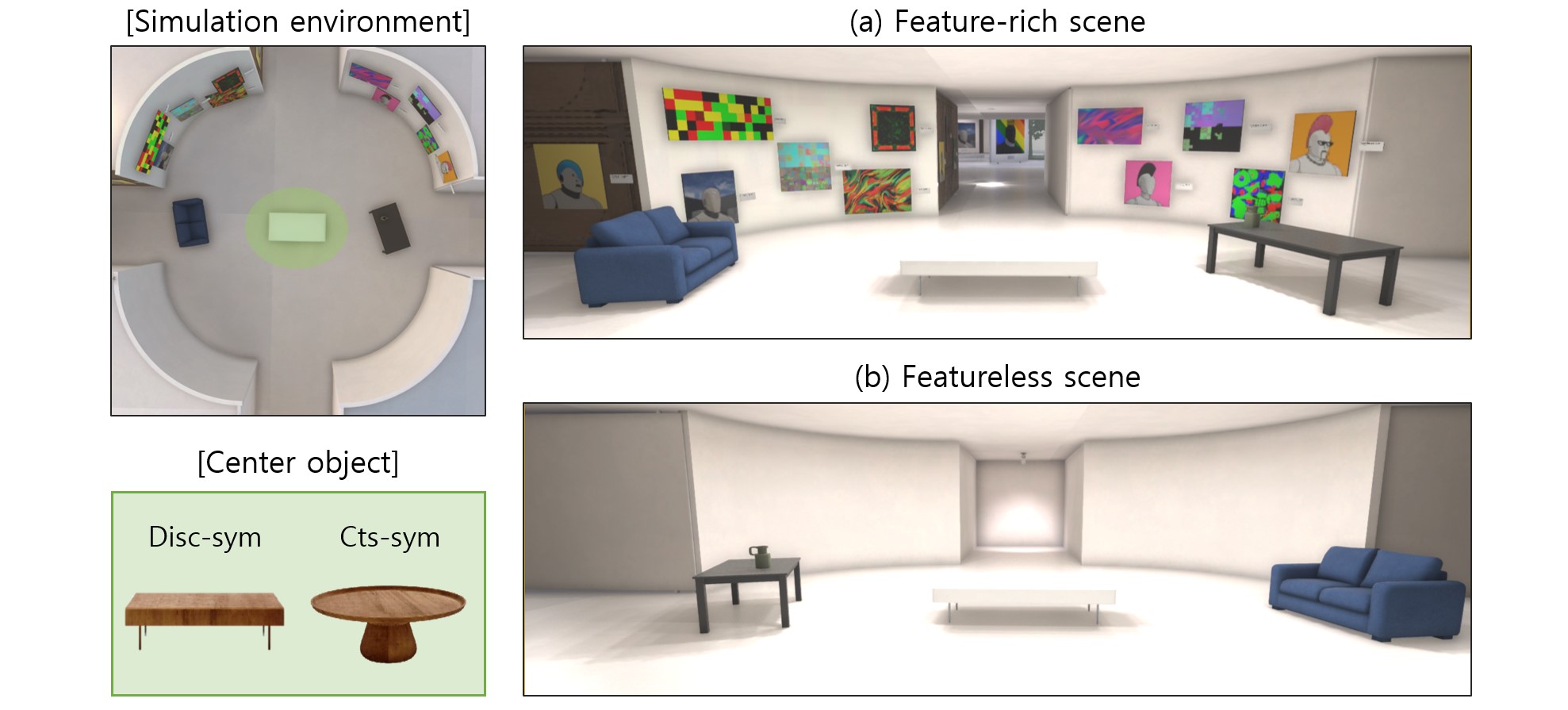}
\caption{\textbf{} Simulation environment.}
\vspace{-2mm}
\label{fig:exp_setup}
\end{figure}

\section{EXPERIMENTAL RESULTS}
\label{experiments and results}
This section presents the performance of the proposed symmetry-aware object SLAM system compared with the baseline. Simulation and public datasets are used to evaluate the proposed system.

\subsection{Setup}
When a camera looks at the feature-rich scene, ego-motion can be estimated accurately by the feature-based SLAM backbone \cite{mur2017orb} in object-based SLAM systems. On the other hand, object-camera constraints are dominant estimation sources if the camera observes a featureless scene.
Therefore, in order to  effectively evaluate the proposed object-based SLAM system, we construct a simulation environment such that one side of the environment has rich feature points and the other side has a few feature points, as shown in Fig. \ref{fig:exp_setup}. Then we obtain the simulation dataset using Unreal Engine and AirSim \cite{shah2018airsim}. The camera moves surrounding a center object in the environment so that camera observes feature-rich and featureless scenes alternately for each specific region, as shown in Fig. \ref{fig:exp_setup} (a) and (b).
As shown in Fig. \ref{fig:exp_setup}, the center object can be replaced with discrete and continuous symmetric objects in the simulation environment. These cases are called \textit{sim(disc)} and \textit{sim(cts)}. In addition, we evaluate the proposed system in the general indoor scenes using the popular scanNet dataset \cite{dai2017scannet}.

For training of the 3D detection network used in our system, pre-training is performed using SUN RGB-D dataset \cite{song2015sun}. After that, to reduce the domain gap between actual and simulation data, additional data are acquired in the simulation, and fine-tuning is performed. We only fine-tune the detection network for simulation dataset since ScanNet dataset only provide axis-aligned bounding box. Both pre-training and fine-tuning is done in the same way as \cite{rupprecht2017learning}. The SUN RGB-D dataset assumes that the input point clouds are represented in the coordinates aligned with the direction of gravity, and expresses the object's orientation using the rotation angle with respect to the axis of gravity (i.e. yaw). 
However, ScanNet dataset has no gravity direction, so we calculated in advance the rotation matrix that aligns the y-axis of camera with the normal vector of the ground plane for each frame. An additional sensor such as an inertial measurement unit or ground detection module may help to change the preprocessing for online implementation. 
We used the number of hypotheses as 30, which was selected empirically to categorize symmetry types of detection well.

We test using  Intel i7-10700 (2.9GHz) and NVIDIA RTX 2060 GPU. DSP-SLAM using single hypothesis detection results is used as the first baseline system (\textit{SH}), and the second baseline (\textit{MH}) is the modified DSP-SLAM to integrate with key idea on \cite{fu2021multi} which uses multi-hypothesis detection with no consideration of symmetry types.

\begin{figure}[t] 
\centering
\includegraphics[width = 1\linewidth]{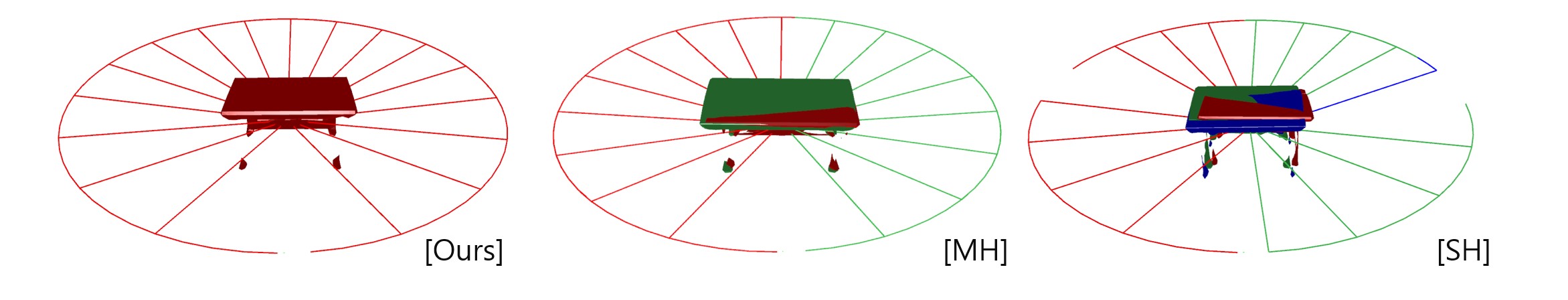}
\caption{\textbf{} The results of object tracking and map object reconstruction in \textit{sim(disc)}.}
\vspace{-5mm}
\label{fig: caseResults_figure}
\end{figure}

\begin{figure}[t] 
\centering
\includegraphics[width = 1\linewidth]{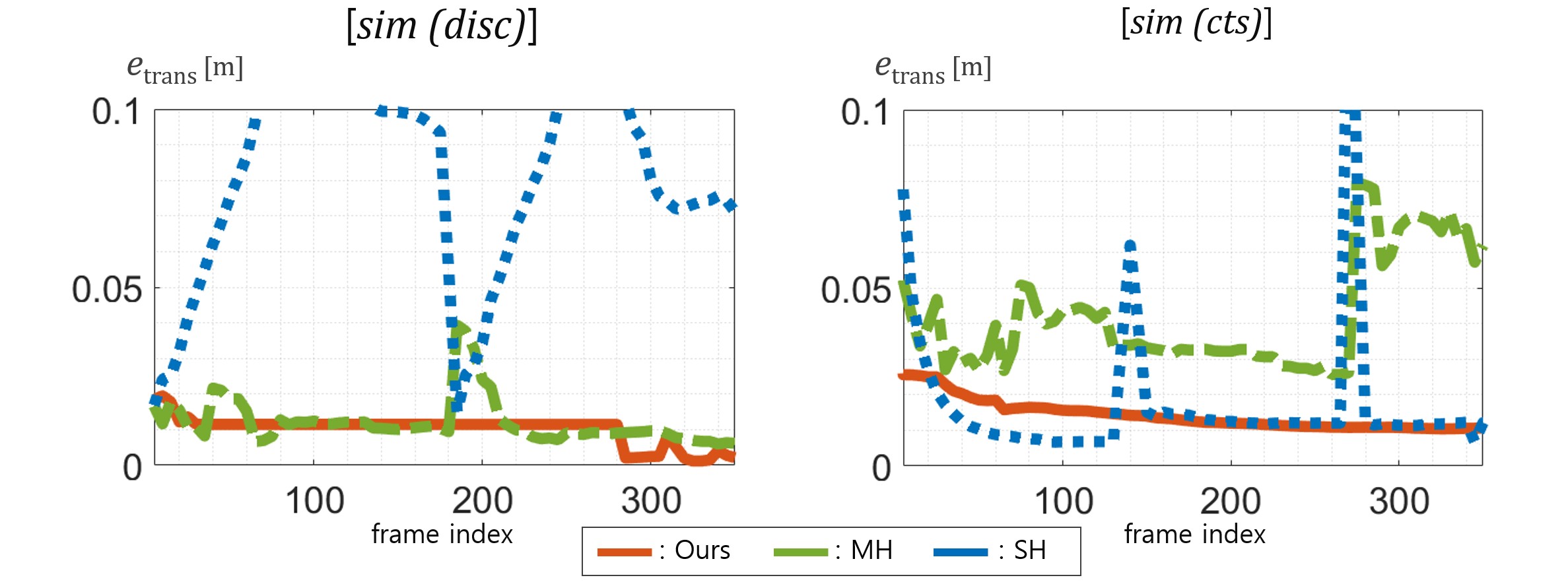}
\caption{\textbf{} The translation error of the estimated map object's pose in case study.}
\vspace{-5mm}
\label{fig: caseResults_graph}
\end{figure}

\begin{figure}[t] 
\centering
\includegraphics[width = 1\linewidth]{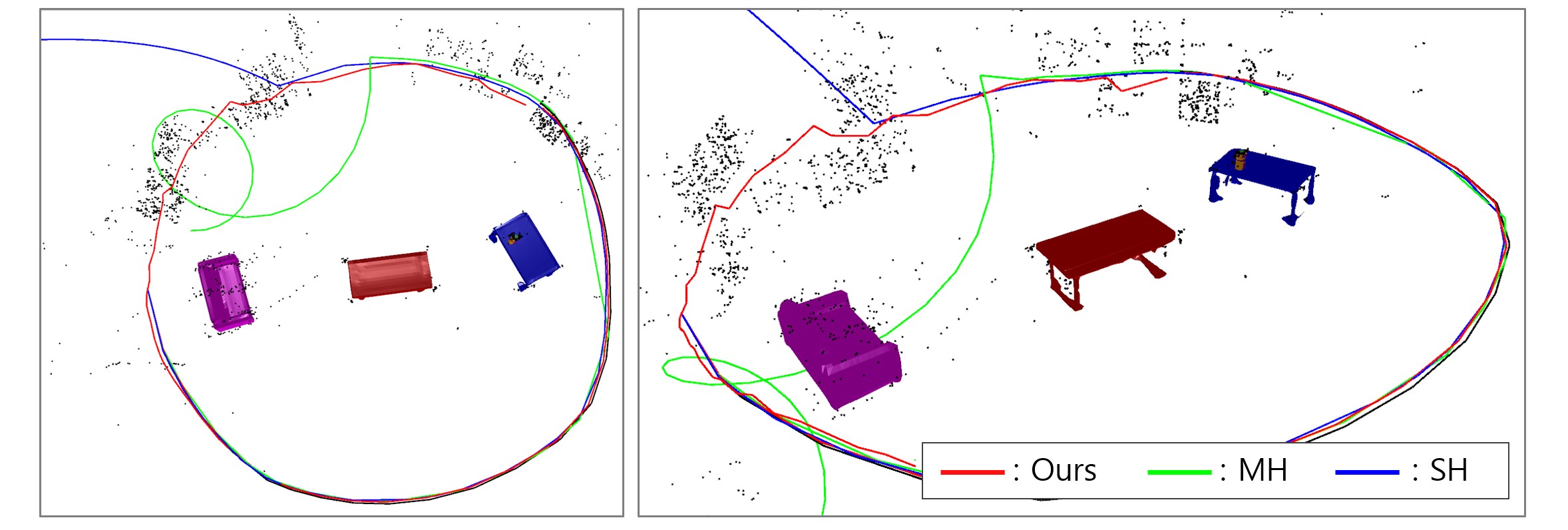}
\caption{\textbf{} The qualitative results of the proposed and baseline algorithms (\textit{SH}, \textit{MH}) in \textit{sim(disc)}.}
\vspace{-2mm}
\label{fig: simResults}
\end{figure}

\begin{figure}[t] 
\centering
\includegraphics[width = 1\linewidth]{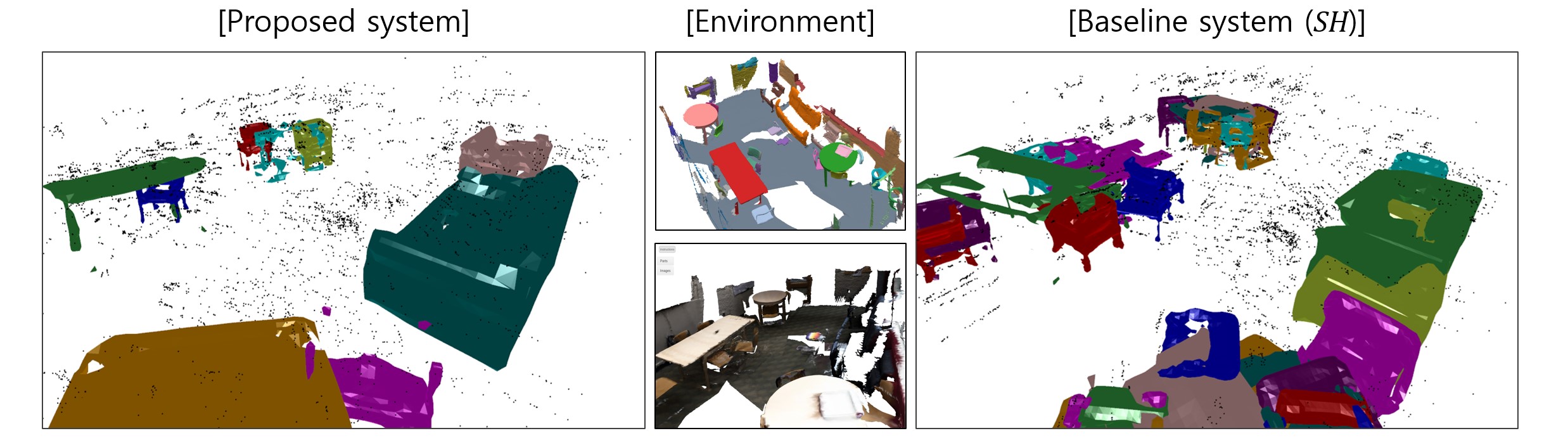}
\caption{\textbf{} The qualitative results of the proposed and baseline (\textit{SH}) algorithms in \textit{Scene0022\_00} in ScanNet.}
\vspace{-4mm}
\label{fig: scanNet_map}
\end{figure}

\subsection{Case study}
To understand how the proposed system enhances the overall performance, we test object tracking and pose estimation under the presence of symmetric objects with pose ambiguity, using the environment in Fig. \ref{fig:exp_setup}. For fair comparison by isolating the other sources that affect the performance except the pose ambiguity, we construct a pose graph using edges from the camera to objects obtained from the proposed categorization and association module and true camera nodes and estimated the object pose by optimization. The same setup is employed  to test the baseline algorithms. 

Fig. \ref{fig: caseResults_figure} shows the result of object tracking and the reconstructed map object using the \textit{sim(disc)} dataset. The proposed method continuously tracks the center object, and a single map object is reconstructed accordingly. On the other hand, the baseline algorithms (\textit{MH}, \textit{SH}) fail to track, and incorrectly recognize a single object as two or more map objects. 

The translation error of the estimated object is reported in Fig. \ref{fig: caseResults_graph}. \textit{MH} and \textit{SH} show large error when establishing the object again after the tracking failure, whereas the proposed method maintains small error through successful object tracking.

\begin{table}[t]
\centering
\caption{\textbf{} The translation error of the proposed and baseline systems using ScanNet and simulation dataset. The bold indicates the best performance and the symbol ‘-’ means failure of pose estimation.}
\label{tab: resultTable}
\begin{tabular}{cc|ccc}
\hline
\multicolumn{2}{c|}{\multirow{2}{*}{dataset}}            & \multicolumn{3}{c}{RMSE of translation {[}m{]}}                                   \\ \cline{3-5} 
\multicolumn{2}{c|}{}                                    & \multicolumn{1}{c|}{OURS}           & \multicolumn{1}{c|}{\textit{SH}}             & \textit{MH}    \\ \hline
\multicolumn{1}{c|}{\multirow{4}{*}{ScanNet}} & \textit{Scene0022\_00} & \multicolumn{1}{c|}{\textbf{0.161}} & \multicolumn{1}{c|}{0.237}          & 0.179 \\
\multicolumn{1}{c|}{}                         & \textit{Scene0049\_00} & \multicolumn{1}{c|}{\textbf{0.137}} & \multicolumn{1}{c|}{\textbf{0.137}} & 0.152 \\
\multicolumn{1}{c|}{}                         & \textit{Scene0091\_00} & \multicolumn{1}{c|}{\textbf{0.079}} & \multicolumn{1}{c|}{0.101}          & 0.091 \\
\multicolumn{1}{c|}{}                         & \textit{Scene0289\_00} & \multicolumn{1}{c|}{\textbf{0.125}}          & \multicolumn{1}{c|}{0.144} & 0.153 \\\hline
\multicolumn{1}{c|}{\multirow{2}{*}{sim}}     & \textit{disc}     & \multicolumn{1}{c|}{\textbf{0.086}} & \multicolumn{1}{c|}{-}              & -     \\
\multicolumn{1}{c|}{}                         & \textit{cts}      & \multicolumn{1}{c|}{\textbf{0.265}} & \multicolumn{1}{c|}{0.341}          & 0.284 \\ \hline
\vspace{-7mm}
\end{tabular}
\end{table}

\subsection{System evaluation}
The performance of the entire system was compared and evaluated with the baseline system using the simulation environment and ScanNet.
First, Fig. \ref{fig: simResults} shows the quantitative results compared with the baseline system using the simulation environment \textit{sim(disc)}. The baseline and camera trajectory are plotted on the map built by the proposed system. In the beginning, a feature-rich scene is observed as shown in Fig. \ref{fig:exp_setup} (a), so all three systems are good at estimating pose. Such trend changes as the features that can be used for localization disappear, since the camera location must be estimated using only the tracked object. Based on the good object tracking performance, the proposed system also demonstrates good performance even in the featureless region using an unambiguous pose parameter that fits the symmetry type. However, both \textit{MH} and \textit{SH} systems fail to estimate the pose in the featureless region. The baseline algorithms associate the detection with the map object on the map using all 6 DoF parameters, and there are cases when none of the multiple hypotheses fits the previous detection result. The data association may fail due to such mismatch in yaw angle. Quantitative results for the simulation environment are shown in Table \ref{tab: resultTable}. For \textit{sim(cts)}, the baseline algorithms also had no failure in pose estimation, but we can see that the proposed algorithm has the highest performance.

Fig. \ref{fig: scanNet_map} shows qualitative results on a ScanNet \textit{Scene0022\_00}. This sequence includes not only the symmetrical objects but also the objects that do not fully enter the camera field of view, so the uncertainty of detection is high.
This is revealed in the result of \textit{SH}. As seen in simulation setting, the baseline system fails data association and many objects are registered in same position of the map. However, the proposed system robustly recognizes as a single object and optimizes the pose  even in these cases. The quantitative results are shown in table \ref{tab: resultTable}. We evaluate the system performance using root mean squared error (RMSE) error of each keyframe position. The proposed system exhibits similar or better path estimation performance in most sequences.

\section{CONCLUSIONS}
\label{conclusions}
Symmetric objects present in the scene  can cause the performance degradation or even failure of SLAM, since their observation at different viewpoints can be identical and cause obscurity. We proposed a method for robustly optimizing the pose of cameras and objects even in the presence of symmetric objects. 

The proposed classification of objects into three categories depending on their symmetry characteristics was successfully applied to various objects. Under the proposed method, the objects in the same category can be associated with the same type of ambiguity, which contributes to the efficiency in data association.
By extracting only the unambiguous parameters corresponding to each category and using them in data association and joint optimization of the camera and object pose, the proposed approach provides significant robustness to the SLAM performance. Proposed system showed better performance than baseline systems in environments with many symmetric objects. 

\addtolength{\textheight}{-8cm}


\bibliographystyle{./bibtex/IEEEtran}
\bibliography{./bibtex/main.bib}

@inproceedings{he2017mask,
  title={Mask r-cnn},
  author={He, Kaiming and Gkioxari, Georgia and Doll{\'a}r, Piotr and Girshick, Ross},
  booktitle={Proceedings of the IEEE international conference on computer vision},
  pages={2961--2969},
  year={2017}
}

@inproceedings{ding2019votenet,
  title={Votenet: A deep learning label fusion method for multi-atlas segmentation},
  author={Ding, Zhipeng and Han, Xu and Niethammer, Marc},
  booktitle={International Conference on Medical Image Computing and Computer-Assisted Intervention},
  pages={202--210},
  year={2019},
  organization={Springer}
}

@article{rukhovich2021fcaf3d,
  title={FCAF3D: Fully Convolutional Anchor-Free 3D Object Detection},
  author={Rukhovich, Danila and Vorontsova, Anna and Konushin, Anton},
  journal={arXiv preprint arXiv:2112.00322},
  year={2021}
}

@article{yang2019cubeslam,
  title={Cubeslam: Monocular 3-d object slam},
  author={Yang, Shichao and Scherer, Sebastian},
  journal={IEEE Transactions on Robotics},
  volume={35},
  number={4},
  pages={925--938},
  year={2019},
  publisher={IEEE}
}

@article{nicholson2018quadricslam,
  title={Quadricslam: Dual quadrics from object detections as landmarks in object-oriented slam},
  author={Nicholson, Lachlan and Milford, Michael and S{\"u}nderhauf, Niko},
  journal={IEEE Robotics and Automation Letters},
  volume={4},
  number={1},
  pages={1--8},
  year={2018},
  publisher={IEEE}
}

@inproceedings{wang2021dsp,
  title={DSP-SLAM: Object oriented SLAM with deep shape priors},
  author={Wang, Jingwen and R{\"u}nz, Martin and Agapito, Lourdes},
  booktitle={2021 International Conference on 3D Vision (3DV)},
  pages={1362--1371},
  year={2021},
  organization={IEEE}
}

@inproceedings{fu2021multi,
  title={A Multi-Hypothesis Approach to Pose Ambiguity in Object-Based SLAM},
  author={Fu, Jiahui and Huang, Qiangqiang and Doherty, Kevin and Wang, Yue and Leonard, John J},
  booktitle={2021 IEEE/RSJ International Conference on Intelligent Robots and Systems (IROS)},
  pages={7639--7646},
  year={2021},
  organization={IEEE}
}


@inproceedings{song2015sun,
  title={Sun rgb-d: A rgb-d scene understanding benchmark suite},
  author={Song, Shuran and Lichtenberg, Samuel P and Xiao, Jianxiong},
  booktitle={Proceedings of the IEEE conference on computer vision and pattern recognition},
  pages={567--576},
  year={2015}
}

@inproceedings{dai2017scannet,
  title={Scannet: Richly-annotated 3d reconstructions of indoor scenes},
  author={Dai, Angela and Chang, Angel X and Savva, Manolis and Halber, Maciej and Funkhouser, Thomas and Nie{\ss}ner, Matthias},
  booktitle={Proceedings of the IEEE conference on computer vision and pattern recognition},
  pages={5828--5839},
  year={2017}
}

@inproceedings{salas2013slam++,
  title={Slam++: Simultaneous localisation and mapping at the level of objects},
  author={Salas-Moreno, Renato F and Newcombe, Richard A and Strasdat, Hauke and Kelly, Paul HJ and Davison, Andrew J},
  booktitle={Proceedings of the IEEE conference on computer vision and pattern recognition},
  pages={1352--1359},
  year={2013}
}

@inproceedings{sucar2020nodeslam,
  title={NodeSLAM: Neural object descriptors for multi-view shape reconstruction},
  author={Sucar, Edgar and Wada, Kentaro and Davison, Andrew},
  booktitle={2020 International Conference on 3D Vision (3DV)},
  pages={949--958},
  year={2020},
  organization={IEEE}
}

@inproceedings{corona2018pose,
  title={Pose estimation for objects with rotational symmetry},
  author={Corona, Enric and Kundu, Kaustav and Fidler, Sanja},
  booktitle={2018 IEEE/RSJ International Conference on Intelligent Robots and Systems (IROS)},
  pages={7215--7222},
  year={2018},
  organization={IEEE}
}

@inproceedings{manhardt2019explaining,
  title={Explaining the ambiguity of object detection and 6d pose from visual data},
  author={Manhardt, Fabian and Arroyo, Diego Martin and Rupprecht, Christian and Busam, Benjamin and Birdal, Tolga and Navab, Nassir and Tombari, Federico},
  booktitle={Proceedings of the IEEE/CVF International Conference on Computer Vision},
  pages={6841--6850},
  year={2019}
}

@article{deng2021poserbpf,
  title={Poserbpf: A rao--blackwellized particle filter for 6-d object pose tracking},
  author={Deng, Xinke and Mousavian, Arsalan and Xiang, Yu and Xia, Fei and Bretl, Timothy and Fox, Dieter},
  journal={IEEE Transactions on Robotics},
  volume={37},
  number={5},
  pages={1328--1342},
  year={2021},
  publisher={IEEE}
}

@inproceedings{merrill2022symmetry,
  title={Symmetry and Uncertainty-Aware Object SLAM for 6DoF Object Pose Estimation},
  author={Merrill, Nathaniel and Guo, Yuliang and Zuo, Xingxing and Huang, Xinyu and Leutenegger, Stefan and Peng, Xi and Ren, Liu and Huang, Guoquan},
  booktitle={Proceedings of the IEEE/CVF Conference on Computer Vision and Pattern Recognition},
  pages={14901--14910},
  year={2022}
}

@inproceedings{zhen2022unified,
  title={Unified representation of geometric primitives for Graph-SLAM optimization using decomposed quadrics},
  author={Zhen, Weikun and Yu, Huai and Hu, Yaoyu and Scherer, Sebastian},
  booktitle={2022 International Conference on Robotics and Automation (ICRA)},
  pages={5636--5642},
  year={2022},
  organization={IEEE}
}

@article{guzman2012multiple,
  title={Multiple choice learning: Learning to produce multiple structured outputs},
  author={Guzman-Rivera, Abner and Batra, Dhruv and Kohli, Pushmeet},
  journal={Advances in neural information processing systems},
  volume={25},
  year={2012}
}

@article{mur2017orb,
  title={Orb-slam2: An open-source slam system for monocular, stereo, and rgb-d cameras},
  author={Mur-Artal, Raul and Tard{\'o}s, Juan D},
  journal={IEEE transactions on robotics},
  volume={33},
  number={5},
  pages={1255--1262},
  year={2017},
  publisher={IEEE}
}

@inproceedings{rupprecht2017learning,
  title={Learning in an uncertain world: Representing ambiguity through multiple hypotheses},
  author={Rupprecht, Christian and Laina, Iro and DiPietro, Robert and Baust, Maximilian and Tombari, Federico and Navab, Nassir and Hager, Gregory D},
  booktitle={Proceedings of the IEEE international conference on computer vision},
  pages={3591--3600},
  year={2017}
}

@inproceedings{grisetti2011g2o,
  title={g2o: A general framework for (hyper) graph optimization},
  author={Grisetti, Giorgio and K{\"u}mmerle, Rainer and Strasdat, Hauke and Konolige, Kurt},
  booktitle={Proceedings of the IEEE international conference on robotics and automation (ICRA), Shanghai, China},
  pages={9--13},
  year={2011}
}

@article{olson2013inference,
  title={Inference on networks of mixtures for robust robot mapping},
  author={Olson, Edwin and Agarwal, Pratik},
  journal={The International Journal of Robotics Research},
  volume={32},
  number={7},
  pages={826--840},
  year={2013},
  publisher={SAGE Publications Sage UK: London, England}
}


@inproceedings{park2019deepsdf,
  title={Deepsdf: Learning continuous signed distance functions for shape representation},
  author={Park, Jeong Joon and Florence, Peter and Straub, Julian and Newcombe, Richard and Lovegrove, Steven},
  booktitle={Proceedings of the IEEE/CVF conference on computer vision and pattern recognition},
  pages={165--174},
  year={2019}
}

@inproceedings{ester1996density,
  title={A density-based algorithm for discovering clusters in large spatial databases with noise.},
  author={Ester, Martin and Kriegel, Hans-Peter and Sander, J{\"o}rg and Xu, Xiaowei and others},
  booktitle={kdd},
  volume={96},
  number={34},
  pages={226--231},
  year={1996}
}

@inproceedings{shah2018airsim,
  title={Airsim: High-fidelity visual and physical simulation for autonomous vehicles},
  author={Shah, Shital and Dey, Debadeepta and Lovett, Chris and Kapoor, Ashish},
  booktitle={Field and service robotics},
  pages={621--635},
  year={2018},
  organization={Springer}
}


@inproceedings{du20203dcfs,
  title={3dcfs: Fast and robust joint 3d semantic-instance segmentation via coupled feature selection},
  author={Du, Liang and Tan, Jingang and Xue, Xiangyang and Chen, Lili and Wen, Hongkai and Feng, Jianfeng and Li, Jiamao and Zhang, Xiaolin},
  booktitle={2020 IEEE International Conference on Robotics and Automation (ICRA)},
  pages={6868--6875},
  year={2020},
  organization={IEEE}
}

@inproceedings{xu2019mid,
  title={Mid-fusion: Octree-based object-level multi-instance dynamic slam},
  author={Xu, Binbin and Li, Wenbin and Tzoumanikas, Dimos and Bloesch, Michael and Davison, Andrew and Leutenegger, Stefan},
  booktitle={2019 International Conference on Robotics and Automation (ICRA)},
  pages={5231--5237},
  year={2019},
  organization={IEEE}
}

@inproceedings{pon2020object,
  title={Object-centric stereo matching for 3d object detection},
  author={Pon, Alex D and Ku, Jason and Li, Chengyao and Waslander, Steven L},
  booktitle={2020 IEEE International Conference on Robotics and Automation (ICRA)},
  pages={8383--8389},
  year={2020},
  organization={IEEE}
}

\begin{thebibliography}{10}
\providecommand{\url}[1]{#1}
\csname url@rmstyle\endcsname
\providecommand{\newblock}{\relax}
\providecommand{\bibinfo}[2]{#2}
\providecommand\BIBentrySTDinterwordspacing{\spaceskip=0pt\relax}
\providecommand\BIBentryALTinterwordstretchfactor{4}
\providecommand\BIBentryALTinterwordspacing{\spaceskip=\fontdimen2\font plus
\BIBentryALTinterwordstretchfactor\fontdimen3\font minus
  \fontdimen4\font\relax}
\providecommand\BIBforeignlanguage[2]{{%
\expandafter\ifx\csname l@#1\endcsname\relax
\typeout{** WARNING: IEEEtran.bst: No hyphenation pattern has been}%
\typeout{** loaded for the language `#1'. Using the pattern for}%
\typeout{** the default language instead.}%
\else
\language=\csname l@#1\endcsname
\fi
#2}}

\bibitem{he2017mask}
K.~He, G.~Gkioxari, P.~Doll{\'a}r, and R.~Girshick, ``Mask r-cnn,'' in
  \emph{Proceedings of the IEEE international conference on computer vision},
  2017, pp. 2961--2969.

\bibitem{du20203dcfs}
L.~Du, J.~Tan, X.~Xue, L.~Chen, H.~Wen, J.~Feng, J.~Li, and X.~Zhang, ``3dcfs:
  Fast and robust joint 3d semantic-instance segmentation via coupled feature
  selection,'' in \emph{2020 IEEE International Conference on Robotics and
  Automation (ICRA)}.\hskip 1em plus 0.5em minus 0.4em\relax IEEE, 2020, pp.
  6868--6875.

\bibitem{xu2019mid}
B.~Xu, W.~Li, D.~Tzoumanikas, M.~Bloesch, A.~Davison, and S.~Leutenegger,
  ``Mid-fusion: Octree-based object-level multi-instance dynamic slam,'' in
  \emph{2019 International Conference on Robotics and Automation (ICRA)}.\hskip
  1em plus 0.5em minus 0.4em\relax IEEE, 2019, pp. 5231--5237.

\bibitem{rukhovich2021fcaf3d}
D.~Rukhovich, A.~Vorontsova, and A.~Konushin, ``Fcaf3d: Fully convolutional
  anchor-free 3d object detection,'' \emph{arXiv preprint arXiv:2112.00322},
  2021.

\bibitem{ding2019votenet}
Z.~Ding, X.~Han, and M.~Niethammer, ``Votenet: A deep learning label fusion
  method for multi-atlas segmentation,'' in \emph{International Conference on
  Medical Image Computing and Computer-Assisted Intervention}.\hskip 1em plus
  0.5em minus 0.4em\relax Springer, 2019, pp. 202--210.

\bibitem{pon2020object}
A.~D. Pon, J.~Ku, C.~Li, and S.~L. Waslander, ``Object-centric stereo matching
  for 3d object detection,'' in \emph{2020 IEEE International Conference on
  Robotics and Automation (ICRA)}.\hskip 1em plus 0.5em minus 0.4em\relax IEEE,
  2020, pp. 8383--8389.

\bibitem{yang2019cubeslam}
S.~Yang and S.~Scherer, ``Cubeslam: Monocular 3-d object slam,'' \emph{IEEE
  Transactions on Robotics}, vol.~35, no.~4, pp. 925--938, 2019.

\bibitem{salas2013slam++}
R.~F. Salas-Moreno, R.~A. Newcombe, H.~Strasdat, P.~H. Kelly, and A.~J.
  Davison, ``Slam++: Simultaneous localisation and mapping at the level of
  objects,'' in \emph{Proceedings of the IEEE conference on computer vision and
  pattern recognition}, 2013, pp. 1352--1359.

\bibitem{fu2021multi}
J.~Fu, Q.~Huang, K.~Doherty, Y.~Wang, and J.~J. Leonard, ``A multi-hypothesis
  approach to pose ambiguity in object-based slam,'' in \emph{2021 IEEE/RSJ
  International Conference on Intelligent Robots and Systems (IROS)}.\hskip 1em
  plus 0.5em minus 0.4em\relax IEEE, 2021, pp. 7639--7646.

\bibitem{nicholson2018quadricslam}
L.~Nicholson, M.~Milford, and N.~S{\"u}nderhauf, ``Quadricslam: Dual quadrics
  from object detections as landmarks in object-oriented slam,'' \emph{IEEE
  Robotics and Automation Letters}, vol.~4, no.~1, pp. 1--8, 2018.

\bibitem{sucar2020nodeslam}
E.~Sucar, K.~Wada, and A.~Davison, ``Nodeslam: Neural object descriptors for
  multi-view shape reconstruction,'' in \emph{2020 International Conference on
  3D Vision (3DV)}.\hskip 1em plus 0.5em minus 0.4em\relax IEEE, 2020, pp.
  949--958.

\bibitem{wang2021dsp}
J.~Wang, M.~R{\"u}nz, and L.~Agapito, ``Dsp-slam: Object oriented slam with
  deep shape priors,'' in \emph{2021 International Conference on 3D Vision
  (3DV)}.\hskip 1em plus 0.5em minus 0.4em\relax IEEE, 2021, pp. 1362--1371.

\bibitem{corona2018pose}
E.~Corona, K.~Kundu, and S.~Fidler, ``Pose estimation for objects with
  rotational symmetry,'' in \emph{2018 IEEE/RSJ International Conference on
  Intelligent Robots and Systems (IROS)}.\hskip 1em plus 0.5em minus
  0.4em\relax IEEE, 2018, pp. 7215--7222.

\bibitem{manhardt2019explaining}
F.~Manhardt, D.~M. Arroyo, C.~Rupprecht, B.~Busam, T.~Birdal, N.~Navab, and
  F.~Tombari, ``Explaining the ambiguity of object detection and 6d pose from
  visual data,'' in \emph{Proceedings of the IEEE/CVF International Conference
  on Computer Vision}, 2019, pp. 6841--6850.

\bibitem{deng2021poserbpf}
X.~Deng, A.~Mousavian, Y.~Xiang, F.~Xia, T.~Bretl, and D.~Fox, ``Poserbpf: A
  rao--blackwellized particle filter for 6-d object pose tracking,'' \emph{IEEE
  Transactions on Robotics}, vol.~37, no.~5, pp. 1328--1342, 2021.

\bibitem{merrill2022symmetry}
N.~Merrill, Y.~Guo, X.~Zuo, X.~Huang, S.~Leutenegger, X.~Peng, L.~Ren, and
  G.~Huang, ``Symmetry and uncertainty-aware object slam for 6dof object pose
  estimation,'' in \emph{Proceedings of the IEEE/CVF Conference on Computer
  Vision and Pattern Recognition}, 2022, pp. 14\,901--14\,910.

\bibitem{zhen2022unified}
W.~Zhen, H.~Yu, Y.~Hu, and S.~Scherer, ``Unified representation of geometric
  primitives for graph-slam optimization using decomposed quadrics,'' in
  \emph{2022 International Conference on Robotics and Automation (ICRA)}.\hskip
  1em plus 0.5em minus 0.4em\relax IEEE, 2022, pp. 5636--5642.

\bibitem{olson2013inference}
E.~Olson and P.~Agarwal, ``Inference on networks of mixtures for robust robot
  mapping,'' \emph{The International Journal of Robotics Research}, vol.~32,
  no.~7, pp. 826--840, 2013.

\bibitem{rupprecht2017learning}
C.~Rupprecht, I.~Laina, R.~DiPietro, M.~Baust, F.~Tombari, N.~Navab, and G.~D.
  Hager, ``Learning in an uncertain world: Representing ambiguity through
  multiple hypotheses,'' in \emph{Proceedings of the IEEE international
  conference on computer vision}, 2017, pp. 3591--3600.

\bibitem{guzman2012multiple}
A.~Guzman-Rivera, D.~Batra, and P.~Kohli, ``Multiple choice learning: Learning
  to produce multiple structured outputs,'' \emph{Advances in neural
  information processing systems}, vol.~25, 2012.

\bibitem{ester1996density}
M.~Ester, H.-P. Kriegel, J.~Sander, X.~Xu, \emph{et~al.}, ``A density-based
  algorithm for discovering clusters in large spatial databases with noise.''
  in \emph{kdd}, vol.~96, no.~34, 1996, pp. 226--231.

\bibitem{park2019deepsdf}
J.~J. Park, P.~Florence, J.~Straub, R.~Newcombe, and S.~Lovegrove, ``Deepsdf:
  Learning continuous signed distance functions for shape representation,'' in
  \emph{Proceedings of the IEEE/CVF conference on computer vision and pattern
  recognition}, 2019, pp. 165--174.

\bibitem{grisetti2011g2o}
G.~Grisetti, R.~K{\"u}mmerle, H.~Strasdat, and K.~Konolige, ``g2o: A general
  framework for (hyper) graph optimization,'' in \emph{Proceedings of the IEEE
  international conference on robotics and automation (ICRA), Shanghai, China},
  2011, pp. 9--13.

\bibitem{mur2017orb}
R.~Mur-Artal and J.~D. Tard{\'o}s, ``Orb-slam2: An open-source slam system for
  monocular, stereo, and rgb-d cameras,'' \emph{IEEE transactions on robotics},
  vol.~33, no.~5, pp. 1255--1262, 2017.

\bibitem{shah2018airsim}
S.~Shah, D.~Dey, C.~Lovett, and A.~Kapoor, ``Airsim: High-fidelity visual and
  physical simulation for autonomous vehicles,'' in \emph{Field and service
  robotics}.\hskip 1em plus 0.5em minus 0.4em\relax Springer, 2018, pp.
  621--635.

\bibitem{dai2017scannet}
A.~Dai, A.~X. Chang, M.~Savva, M.~Halber, T.~Funkhouser, and M.~Nie{\ss}ner,
  ``Scannet: Richly-annotated 3d reconstructions of indoor scenes,'' in
  \emph{Proceedings of the IEEE conference on computer vision and pattern
  recognition}, 2017, pp. 5828--5839.

\bibitem{song2015sun}
S.~Song, S.~P. Lichtenberg, and J.~Xiao, ``Sun rgb-d: A rgb-d scene
  understanding benchmark suite,'' in \emph{Proceedings of the IEEE conference
  on computer vision and pattern recognition}, 2015, pp. 567--576.

\end{thebibliography}

\end{document}